\documentclass{article}

\usepackage{microtype}
\usepackage{graphicx}
\usepackage{subfigure}
\usepackage{booktabs} 

\usepackage{hyperref}

\newcommand*\diff{\mathop{}\!\mathrm{d}}

\usepackage[accepted]{icml2023}

\usepackage{amsmath}
\usepackage{amssymb}
\usepackage{mathtools}
\usepackage{amsthm}

\usepackage[capitalize,noabbrev]{cleveref}

\theoremstyle{plain}
\newtheorem{theorem}{Theorem}[section]
\newtheorem{proposition}[theorem]{Proposition}
\newtheorem{lemma}[theorem]{Lemma}

\theoremstyle{definition}

\theoremstyle{remark}

\usepackage[textsize=tiny]{todonotes}

\definecolor{myblue}{rgb}{0.87,0.92,0.96}
\usepackage{multirow}
\usepackage{stfloats}
\usepackage{algorithmic}
\usepackage{algorithm}
\newcommand{\argmax}[1] {{\operatorname{argmax}}_{#1}}
\usepackage{array}

\icmltitlerunning{Boosting Offline Reinforcement Learning with Action Preference Query}

\begin{document}

\twocolumn[
\icmltitle{Boosting Offline Reinforcement Learning with Action Preference Query}

\icmlsetsymbol{equal}{*}

\begin{icmlauthorlist}
\icmlauthor{Qisen Yang}{equal,sch}
\icmlauthor{Shenzhi Wang}{equal,sch}
\icmlauthor{Matthieu Gaetan Lin}{sch2}
\icmlauthor{Shiji Song}{sch}
\icmlauthor{Gao Huang}{sch}
\end{icmlauthorlist}

\icmlaffiliation{sch2}{Department of Computer Science, BNRist, Tsinghua University, Beijing, China}
\icmlaffiliation{sch}{Department of Automation, BNRist, Tsinghua University, Beijing, China}

\icmlcorrespondingauthor{Gao Huang}{gaohuang@tsinghua.edu.cn}

\icmlkeywords{Machine Learning, ICML}

\vskip 0.3in
]

\printAffiliationsAndNotice{\icmlEqualContribution} 

\begin{abstract}
    Training practical agents usually involve offline and online reinforcement learning (RL) to balance the policy's performance and interaction costs.
In particular, online fine-tuning has become a commonly used method to correct the erroneous estimates of out-of-distribution data learned in the offline training phase.
However, even limited online interactions can be inaccessible or catastrophic for high-stake scenarios like healthcare and autonomous driving.
In this work, we introduce an interaction-free training scheme dubbed Offline-with-Action-Preferences (OAP).
The main insight is that, compared to online fine-tuning, querying the preferences between pre-collected and learned actions can be equally or even more helpful to the erroneous estimate problem.
By adaptively~encouraging or suppressing policy constraint according to action preferences, OAP could distinguish overestimation from beneficial policy improvement and thus attains a more accurate evaluation of unseen data.
Theoretically, we prove a lower bound of the behavior policy's performance improvement brought by OAP.
Moreover, comprehensive experiments on the D4RL benchmark and state-of-the-art algorithms demonstrate that OAP yields higher (29\% on average) scores, especially on challenging AntMaze tasks (98\% higher).
\end{abstract}

\section{Introduction}
Traditionally, reinforcement learning (RL) algorithms iteratively collect experience by interacting with the environment~\cite{sutton1998introduction}.
However, such interactions are impractical in many applications, either because online data collection is expensive or dangerous \cite{yue2022value}, \textit{e.g.} in robotics\cite{robotics_survey}, educational agents~\cite{education_survey}, healthcare~\cite{healthcare_survey}, and self-driving~\cite{autonomous_driving_survey}.
Therefore, offline RL, where agents only learn from a pre-collected dataset without any additional online interaction, emerges and thrives~\cite{offline_survey_sergey}.

\begin{figure}[ht]
\centering
    \includegraphics[width=0.48\textwidth]{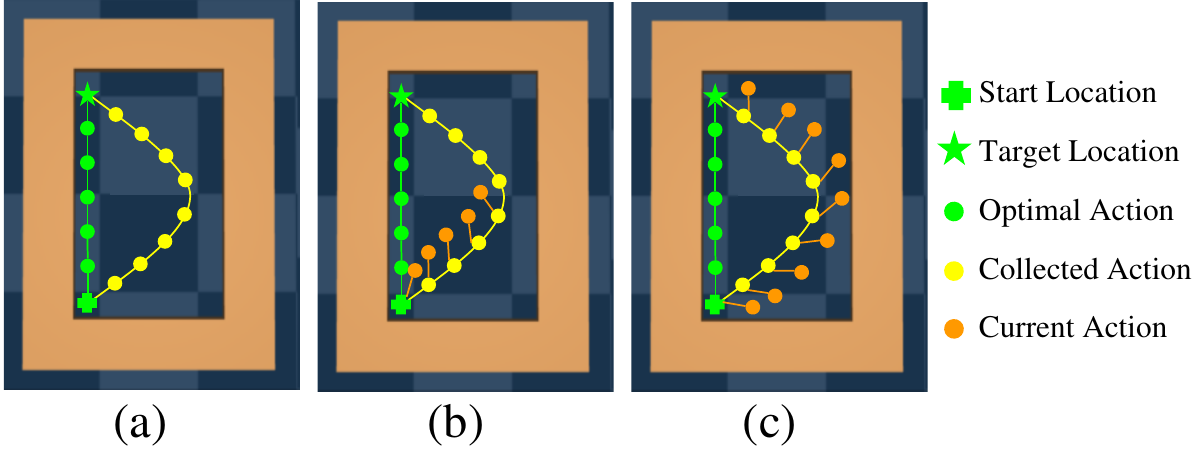}
    \vskip -0.02 in
    \caption{An intuitive example of action preferences. (a) The task is to find the shortest path. (b) If the current policy acts better than the pre-collected action, it shows that the current learning direction is right and should be supported. (c) Otherwise, the OOD data are overestimated, and the distributional shift should be~suppressed.}
    \label{fig: intuitive example}
\end{figure}

Moreover, learning from a static dataset also makes offline RL suffer from the \textit{distributional shift}, \textit{i.e.}, the gap between state-action distributions of the training data and the test environment.
It hinders the agent's performance due to off-policy bootstrapping error accumulation caused by out-of-distribution (OOD) data, \textit{i.e.}, data that is out of the distribution of the offline dataset.
This problem is inevitable because there exists a counterfactual inference problem: given data that resulted from a set of decisions, infer the consequence of a different set of decisions~\cite{offline_survey_sergey}.
Hence, offline RL agents are usually fine-tuned by further online training~\cite{AWAC,IQL}, where erroneous estimates of OOD data could be corrected through accurate rewards and real-time transitions.
Notably, unlike regular online algorithms, the online fine-tuning phase usually has a limited interaction budget.

Nevertheless, for high-risk scenarios like healthcare and autonomous driving, even a few online interactions with the environment can cause catastrophic losses.
Additionally, designing an appropriate reward function takes significant effort for many real-world environments.
In this case, developing a safer approach to boosting offline RL without any online interactions is valuable.
Supposing that there is a measure where the optimality of given actions can be queried, it may be equally or even more helpful to the erroneous estimate problem if we adaptively support or suppress the agent to learn unseen data in the training process, compared to the online fine-tuning approach.
As in \cref{fig: intuitive example}, the task is to find the optimal path to the target location from the start point.
There are pre-collected actions in the static offline dataset.
The agent may learn OOD actions because of the policy improvement objective.
Intuitively, if OOD actions are better than the collected ones, the current learning direction is encouraged.
Conversely, if they are worse than the collected ones, we suppress the divergence and urge the agent to imitate the offline dataset.

This paper proposes  a novel query-based offline training scheme, dubbed Offline-with-Action-Preferences (OAP), to achieve the adaptive constraint.
Unlike the commonly used interaction-based method, we periodically query preferences between pre-collected and learned actions during offline training and adjust optimization directions according to these action preferences.
Instead of high-performing demonstrations, acquiring action preferences are viable in real-world deployment because available expert models are usually proprietary and can only be accessed in an interaction-free and blackbox way~\cite{yu2020dynamic,chi2021tohan}.
Specifically, OAP involves three steps: (1) optionally selecting samples in the offline dataset and querying for preferences, (2) learning the preference pattern using a neural network and pseudo-annotating the rest of the data, (3) training the agent with the adjusted optimization objective.
Theoretically, we prove that OAP brings a stable performance improvement of the behavior policy, and even inaccurate preference annotations could help.
Empirically, we instantiate OAP with state-of-the-art offline RL algorithms and perform proof-of-concept investigations on the D4RL benchmark~\cite{Fu2020D4RLDF}.
Surprisingly, compared to online fine-tuning, OAP is safer because online interactions are not required, leading to significantly higher scores, especially on challenging tasks (up to 115\%~higher).

\section{Preliminaries}
\begin{figure*}[ht]
\centering
    \includegraphics[width=0.99\textwidth]{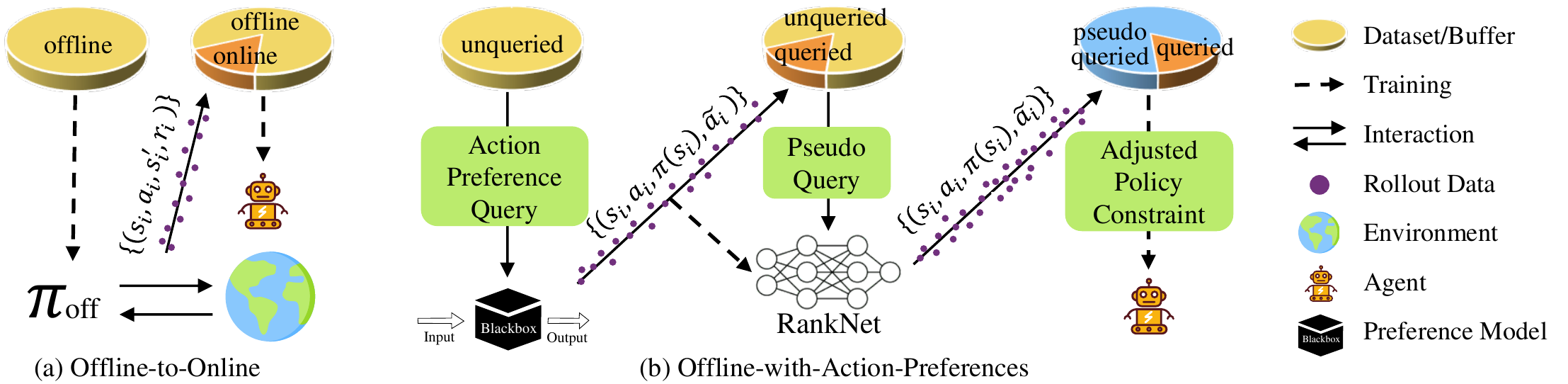}
    \vskip -0.02 in
    \caption{Schematic illustrations of the commonly-used Offline-to-Online and the proposed OAP paradigms. \textbf{(a)} The agent learns a policy $\pi_{\mathrm{off}}$ from the pre-collected offline dataset and collects new experiences by interacting with the environment to further fine-tune the policy. \textbf{(b)} Some samples in the offline dataset are selected and annotated with action preferences by the proprietary preference model in a blackbox way. Then a RankNet learns from the queried samples and conducts pseudo queries on the rest of the data. All the annotated data are used to train the offline agent with the adjusted updating objective. The entire process does not involve any online interactions.}
    \label{fig: method}
\end{figure*}

\textbf{RL formulation.} RL tasks are usually modeled as a Markov decision process (MDP) which can be denoted as a tuple $\mathcal{M}=(\mathcal{S}, \mathcal{A}, T, \rho^0, R, \gamma)$.
$\mathcal{S}$ is the state space, $\mathcal{A}$ is the action space, $T(s_{t+1}|s_t,a_t)$ defines the transition function of the environment $E$, $\rho^0(s_0)$ is the initial state distribution, $R:\mathcal{S}\times\mathcal{A}\rightarrow\mathbb{R}$ defines the reward function, and $\gamma\in(0,1]$ is a scalar discount factor.
The objective of RL is to learn a policy $\pi(a_t|s_t)$ that maximizes the accumulated rewards.

\textbf{Offline RL.} Offline RL can be seen as a data-driven formulation of the reinforcement learning problem.
The agent cannot interact with the environment and only learns from a previously collected dataset $\mathcal{D} = \{(s_i, a_i, s^{\prime}_i, r_i) \mid i=1, 2, \cdots, N\}$.
The distribution over states and actions in $\mathcal{D}$ is denoted as the behavior policy  $\pi_{\beta}$.
Finding a balance between increased generalization and avoiding unwanted
behaviors outside of distribution is one of the core problems of offline RL~\cite{offline_survey_rafael}.
Generally speaking, the optimization objectives of popular offline RL algorithms~\cite{CQL,AWAC,td3+bc,IQL} make trade-offs between policy improvement and policy constraint, either explicitly or implicitly.
It can be formulated as follows:
\begin{equation}
\label{eq: unified objective}
    \begin{aligned}
    \pi^* =& \argmax{\pi} \mathbb{E}_{(s, a)\sim \mathcal{D}} \operatorname{F} \Big(\underbrace{L_{pi}\left(Q, \pi, s, a\right)}_{\text{policy improvement term}}, \\ 
    & \underbrace{L_{pc}\left(Q, \pi, s, a\right)}_{\text{policy constraint term}}, \underbrace{d_c}_{\text{constraint degree}}\Big), 
    \end{aligned}
\end{equation}

\vspace{-8pt}
where $\pi$ is a policy, $\pi^*$ is the optimal policy, $Q(s, a): \mathcal{S}\times \mathcal{A}\to \mathbb{R}$ is a state-action value function estimating the expected sum of discounted rewards after taking action $a$ at state $s$.~\cite{sutton1998introduction,offline_survey_sergey}

\section{Method}
\label{sec: method}
The commonly-used Offline-to-Online scheme is based on the pretrain-finetune paradigm, as shown in \cref{fig: method}(a). 
A policy $\pi_{\mathrm{off}}$ is first learned from the offline dataset that stores pre-collected experiences.
This process does not involve online interactions and is thus friendly to high-stake applications.
Considering the erroneous estimate problem caused by learning from a static dataset, $\pi_{\mathrm{off}}$ is further fine-tuned by a few online interactions with the environment.
The second online process can achieve significant performance improvement for the agent but also brings high risk and cost.
We aim to propose a new training scheme that exempts the agent from online interactions and meanwhile improves the offline policy $\pi_{\mathrm{off}}$.
As shown in \cref{fig: method}(b) and Algorithm \ref{alg: algos-OAP}, our OAP scheme first queries a few samples for action preferences (\cref{sec: method Action Preference Query}).
Secondly, the rest of the unqueried data are pseudo-queried by a learned RankNet (\cref{sec: method Pseudo Query with RankNet}).
Thirdly, these queried and pseudo-queried data are used to train the agent with the adjusted policy constraint (\cref{sec: method Adjusted Policy Constraint}).
The three main elements and further theoretical analyses of OAP are detailedly described below.

\subsection{Action Preference Query}
\label{sec: method Action Preference Query}

Offline RL aims to optimize the policy $\pi$ by an offline dataset $\mathcal{D} = \{(s_i, a_i, s^{\prime}_i, r_i) \mid i=1, 2, \cdots, N\}$.
As in preference-based RL, an action preference compares two actions for the same state~\cite{preference_survey}.
Given a state-action pair $(s_i, a_i)\in\mathcal{D}$ and a preference function $G$, the action preference query can be formulated as $\tilde{a}_i = G(s_i, a_i, \pi(s_i))$, where $\pi$ is the current policy and $\tilde{a}_i$ is the preferred action. 
In real-world applications, the available preference model for queries is usually proprietary, and the preferred action can be annotated in a blackbox way.
In this paper, we train an expert policy beforehand and utilize its state-action value function $Q^*(s,a)$ end-to-end to serve as the proprietary model. The preference function is:
\begin{equation}
    \label{eq: pr_true}
    G(s_i, a_i, \pi(s_i)) = \mathop{\arg\max}\limits_{a\in\{a_i,\pi(s_i)\}}Q^*(s_i,a).
\end{equation}

\vspace{-8pt}
For fair comparisons, the number of queries is limited to 100k as the interaction steps in the Offline-to-Online scheme, which is usually accessible and economical in real-world scenarios.
Therefore, only the most divergent actions, which may suffer more from the distributional shift problem, are considered worthy of being queried.
In other words, millions of samples in the dataset are ranked according to the divergence criterion, and the top ones are selected for action preferences.
We simply adopt the Euclidean norm as the \textit{ranking criterion}: $l_i = (\pi(s_i)-a_i)^2$.

\subsection{Pseudo Query with RankNet}
\label{sec: method Pseudo Query with RankNet}

To take full advantage of query information, all queried samples are collected as a query dataset $\mathcal{D}_q = \{(s_k, a_k,\pi^k(s_k),\tilde{a}_k) \mid k=1, 2, \cdots, M\}$. 
Meanwhile, the action preference problem can be viewed as ranking two options under the same state.
Considering that practical recommendation systems can learn a ranking function based on a few query pairs, we attempt to obtain pseudo query results by learning from the query dataset $\mathcal{D}_q$.
One of the typical Learning-to-Rank approaches, RankNet~\cite{ranknet}, is adopted in our method.
It models the underlying ranking function $f_r$ by a neural network with 3 MLP layers.
Denote the modeled posterior $P(f_r(s_k,a_k)>f_r(s_k,\pi^k(s_k)))$ by $P_{k}, k=1, 2, \cdots, M$, and let $\Bar{P}_{k}$ be the logged target values for those posteriors.
Define $o_{k}=f_r(s_k,a_k)-f_r(s_k,\pi^k(s_k)))$.
The pairwise cost function of RankNet is formulated as follows:
\begin{equation}
    \label{eq: cost function of ranknet_1}
    C_{k}=C(o_{k})=-\Bar{P}_{k}\mathrm{log}P_{k}-(1-\Bar{P}_{k})\mathrm{log}(1-P_{k}),
\end{equation}

where the map from outputs to probabilities is modeled using a logistic function $P_k = e^{o_k}/(1+e^{o_k})$.

After querying the selected samples and training RankNet by \cref{eq: cost function of ranknet_1}, the rest of the samples in the offline dataset are pseudo-queried by RankNet.
Then the preference function in this process is changed to:
\begin{equation}
    \label{eq: pr_ranknet}
    G(s_i, a_i, \pi(s_i)) = \mathop{\arg\max}\limits_{a\in\{a_i,\pi(s_i)\}}f_r(s_i,a).
\end{equation}

\subsection{Adjusted Policy Constraint}
\label{sec: method Adjusted Policy Constraint}

Offline RL requires reconciling two conflicting aims \cite{IQL}: policy improvement and policy constraint.
Generally, a tight constraint would hinder the policy from improving over the behavior policy that collected the offline dataset.
A loose constraint may result in a policy that suffers from distributional shift~\cite{offline_survey_sergey} and fails on OOD states.
Therefore, we aim to achieve a better policy constraint by adaptively deviating from the fixed dataset based on query results.
Specifically, when the current policy acts better than the pre-collected action, it is unnecessary to exert a strong constraint.
On the contrary, if the action conducted by the current policy is worse than that in the dataset, the policy would remain constrained near the behavior policy.
Taking TD3+BC~\cite{td3+bc} as an example, the original training objective is:
\begin{equation}
    \label{eq: td3+bc}
    \pi=\mathop{\arg\max}\limits_{\pi}\mathbb{E}_{(s,a)}\Big[\underbrace{\lambda Q(s,\pi(s))}_{\text{policy improvement}}-\underbrace{(\pi(s)-a)^2}_{\text{policy constraint}}\Big],
\end{equation}

where $\lambda$ is a hyperparameter and $Q$ is the state-value function.
After the action preference is acquired, the policy constraint term is adjusted, and the objective becomes:
\begin{equation}
    \label{eq: ajusted td3+bc}
    \pi=\mathop{\arg\max}\limits_{\pi}\mathbb{E}_{(s,a)}\left[\lambda Q(s,\pi(s))-(\pi(s)-\tilde{a})^2\right],
\end{equation}

where $\tilde{a}$ refers to the preferred action in \cref{eq: pr_true} and \cref{eq: pr_ranknet}.
Consequently, if the updating direction is correct, then the constraint is loosened near the current policy; but if the direction induces worse actions, the constraint reduces to the behavior policy.
Such adaptive policy constraints would encourage more aggressive policy improvement and meanwhile filter out wrong~moves.

\subsection{Theoretical Analysis}
\label{sec: method Theoretical Analysis}
In this section, we theoretically validate the superiority of OAP.
Considering the blackbox policy that provides action preferences is optimal, \cref{prop: perfect case} suggests that the trained policy is constrained to a better behavior policy. 
Meanwhile, suppose the preferences are faulty because of the proprietary preference model or the learned RankNet. In that case, \cref{prop: imperfect case} demonstrates that OAP can still bring performance improvement.

For any deterministic policy $\pi$, its performance (return) can be formulated as $\eta({\pi}) = \mathbb{E}_{\tau \sim \pi}\left[ \sum_{t=0}^{+\infty} \gamma^t r(s_t, a_t) \right]$.
Denote the behavior policy of the pre-collected offline dataset $\mathcal{D}$ as $\pi_\beta$, and the behavior policy revised with action preferences as $\Tilde{\pi}_\beta$.
For any policy $\pi$, $\rho_{\pi}$ is the (unnormalized)
discounted visitation frequency, defined as $\rho_\pi(s)= \sum_{t=0}^\infty \gamma^t P\left(s_t=s\right)$,
where $s_0\sim \rho^0(s_0)$ and the trajectory $(s_0, s_1, \ldots)$ is sampled by the policy $\pi$. 
By the definition, $\rho_\pi(s) \in [0, \frac{1}{1-\gamma}]$.

\begin{proposition}[Perfect preference case] \label{prop: perfect case}
Consider the case with perfect preferences, \textit{i.e.}, $\forall (s, a)$, the state-action value function $Q^*(s, a)$ used for the action preference query is accurate.
Then $\pi_\beta$ and $\Tilde{\pi}_\beta$ satisfy:
\begin{equation}
\begin{aligned}
    \eta(\Tilde{\pi}_\beta) &- \eta(\pi_\beta) \\
    & \approx \mathbb{E}_{s\sim\mathcal{D}} \left[ Q^*(s, \Tilde{\pi}_\beta(s)) - Q^*(s, \pi_\beta(s)) \right] \ge 0. \label{eq: approximation in the perfect case}
\end{aligned}
\end{equation}
\end{proposition}
\begin{proof}
The proof is deferred to \cref{appendix: the proof of the perfect case}.
\end{proof}

For value functions $Q_1, Q_2$, we define the total variation distance $D_{\mathrm{TV}}^{\pi}(Q_1, Q_2) = \max_{s}|Q_1(s, \pi(s)) - Q_2(s, \pi(s))|$.

\begin{proposition}[Imperfect preference case] \label{prop: imperfect case}
Consider the case where preferences probably have errors.
Denote the accurate state-action value function as $Q^*(s, a)$ and the faulty function as $\Hat{Q}^*(s, a)$.
Then $\forall \hat{Q}^*$ satisfying $ D_{\mathrm{TV}}^{\Tilde{\pi}_\beta}(\Hat{Q}^*, Q^*) \le \Tilde{\alpha}, D_{\mathrm{TV}}^{\pi_\beta}(\Hat{Q}^*, Q^*) \le \alpha$, it holds that
\begin{equation}
\begin{aligned}
    \eta(\Tilde{\pi}_\beta) & - \eta(\pi_\beta) \gtrsim \mathbb{E}_{s\sim \mathcal{D}} \Big[\hat{Q}^*(s, \Tilde{\pi}_\beta(s)) \\
    & -\hat{Q}^*(s, \pi_\beta(s))\Big] - 2 (\Tilde{\alpha} + \alpha) \overline{\rho}_{\pi_\beta},
    \label{eq: imperfect case lower bound}
\end{aligned}
\end{equation}
where $\overline{\rho}_{\pi_\beta} = \sup \{\rho_{\pi_\beta}(s), s\in\mathcal{S}\} \in \left[\frac{1}{|\mathcal{S}_{\mathcal{D}}|(1-\gamma)}, \frac{1}{1-\gamma}\right]$ ($|\mathcal{S}_{\mathcal{D}}|$ denotes the number of different states in $\mathcal{D}$). 
\end{proposition}
\begin{proof}
    Please refer to \cref{appendix: the proof of the imperfect case}.
\end{proof}

The first term in the RHS of \cref{eq: imperfect case lower bound} is non-negative because $\forall s\in \mathcal{S}, \hat{Q}^*(s, \Tilde{\pi}_\beta(s)) - \hat{Q}^*(s, \pi_\beta(s)) \ge 0$ according to \cref{eq: pr_true}.
The second term $-2(\Tilde{\alpha}+\alpha)\overline{\rho}_{\pi_\beta}$ relates to the quality of the blackbox policy.
A more accurate blackbox policy would lead to smaller error bounds ($\alpha, \overline{\alpha}$) and thus a larger performance lower bound of $\Tilde{\pi}_\beta$.
Furthermore, in offline RL, the offline dataset usually contains a large number of samples with various states.
Therefore, $\Tilde{\rho}_{\pi_\beta}$ would be small and close to its lower bound $\frac{1}{|\mathcal{S}_{\mathcal{D}}|(1-\gamma)} \ll 1$.
It indicates that OAP has a tolerance for faulty annotations, and coarse preference results can improve performance when a variety of offline samples are available.

\begin{algorithm}
\caption{Offline-with-Action-Preferences} \label{alg: algos-OAP}
\begin{algorithmic}
    \REQUIRE Offline dataset $\mathcal{D}$, query dataset $\mathcal{D}_q$, training steps $N_\mathrm{train}$, query intervals $M_\mathrm{inter}$, query limit $K_\mathrm{total}$.
    \ENSURE Policy $\pi$ after optimization.
    \STATE Initialize policy $\pi$ and RankNet $f_r$.
    \STATE Let the preferred actions $\tilde{a}_i=a_i, a_i\in\mathcal{D}$.
    \FOR{$t = 1 \to N_\mathrm{train}$}
        \STATE Update the policy $\pi$ by \cref{eq: ajusted td3+bc}.
            \IF{$t\mod M_\mathrm{inter}=0$}
            \STATE Select $\frac{K_\mathrm{total}M_\mathrm{inter}}{N_\mathrm{train}}$ samples from the offline dataset $\mathcal{D}$ according to the ranking criterion.
            \STATE Conduct action preference query by \cref{eq: pr_true}.
            \STATE Add queried samples into the query dataset $\mathcal{D}_q$.
            \FOR{epoch = 0, 1, $\cdots$, until convergence}
            \STATE Train the RankNet $f_r$ with the query dataset $\mathcal{D}_q$ by \cref{eq: cost function of ranknet_1}.
            \ENDFOR
            \STATE Conduct pseudo queries on the rest of the samples in the offline dataset $\mathcal{D}$ by \cref{eq: pr_ranknet}.
            \ENDIF
    \ENDFOR
\end{algorithmic}
\end{algorithm}

\section{Experiments}
\begin{table*}[ht]
  \small
  \caption{Comparisons of five training schemes.
  }
  \centering
  \begin{tabular}{lccccc}
    \specialrule{0.12em}{0pt}{0pt}
	Requirement
	& Offline 
	& Online 
	& Online-Mix 
	& Offline-to-Online
	& OAP
	\\ \hline
    Pre-collected Offline Data 
    & $\checkmark$  &   & $\checkmark$
    & $\checkmark$  & $\checkmark$  
    \\
    Training on Offline Data 
    & $\checkmark$  &   & 
    & $\checkmark$  &   
    \\
    \hline
    Available State Transition 
    &   & $\checkmark$  & $\checkmark$
    & $\checkmark$  &   
    \\
    Predefined Reward Function 
    &  & $\checkmark$  & $\checkmark$
    & $\checkmark$  &   
    \\
    \hline
    Action Preference Query 
    &   &   & 
    &   & $\checkmark$  
    \\
    \specialrule{0.12em}{0pt}{0pt}
  \end{tabular}
  \label{table: setting_comparison}
\end{table*}

\begin{table*}[ht]
  \small
  \caption{Average normalized D4RL score~\cite{Fu2020D4RLDF} over the final 10 evaluations and 5 random seeds. 
  Different training schemes are instantiated on the commonly-used TD3+BC~\cite{td3+bc} algorithm. OAP is safer than the popular Offline-to-Online scheme but performs significantly better on a variety of tasks. The standard error of AntMaze is usually large since the return is binomial.
  }
  \centering
  \begin{tabular}{lccccc}
    \specialrule{0.12em}{0pt}{0pt}
	Dataset
	& Offline 
	& Online 
	& Online-Mix 
	& Offline-to-Online
	& OAP
	\\ \hline
    halfcheetah-random-v2 
    & 11.1 $\pm$ 1.3  & 19.2 $\pm$ 4.1  & 28.3 $\pm$ 2.1
    & \colorbox{myblue}{32.3} $\pm$ 1.9  & 24.0 $\pm$ 1.6  
    \\
    hopper-random-v2 
    & 8.7 $\pm$ 1.6  & \colorbox{myblue}{14.5} $\pm$ 12.5  & 10.1 $\pm$ 0.4
    & 8.6 $\pm$ 1.2  & 8.8 $\pm$ 1.8 
    \\
    walker2d-random-v2 
    & 1.8 $\pm$ 1.5  & 1.0 $\pm$ 2.5  & 2.2 $\pm$ 1.5
    & \colorbox{myblue}{7.7} $\pm$ 8.0  & 5.1 $\pm$ 5.1 
    \\
    halfcheetah-medium-v2 
    & 48.1 $\pm$ 0.2  & 19.2 $\pm$ 4.1  & 48.1 $\pm$ 0.2  & 49.2 $\pm$ 0.2  & \colorbox{myblue}{56.4} $\pm$ 4.3   
    \\ 
    hopper-medium-v2 
    & 55.8 $\pm$ 1.2  & 14.5 $\pm$ 12.5  & 58.4 $\pm$ 1.7  & 57.8 $\pm$ 2.0  & \colorbox{myblue}{82.0} $\pm$ 6.6
    \\ 
    walker2d-medium-v2 
    & 83.2 $\pm$ 0.5  & 1.0 $\pm$ 2.5  & 79.2 $\pm$ 9.9  & 85.1 $\pm$ 0.9  & \colorbox{myblue}{85.6} $\pm$ 1.2   
    \\ 
    halfcheetah-medium-replay-v2 
    & 44.9 $\pm$ 0.1  & 19.2 $\pm$ 4.1  & 46.0 $\pm$ 0.4  & 48.3 $\pm$ 0.3  & \colorbox{myblue}{53.4} $\pm$ 1.9   \\ 
    hopper-medium-replay-v2 
    & 57.2 $\pm$ 10.2  & 14.5 $\pm$ 12.5  & 48.4 $\pm$ 3.4  & 76.3 $\pm$ 13.3  & \colorbox{myblue}{98.5} $\pm$ 2.5   \\ 
    walker2d-medium-replay-v2 
    & 81.1 $\pm$ 2.8  & 1.0 $\pm$ 2.5  & 76.7 $\pm$ 14.1  & \colorbox{myblue}{85.6} $\pm$ 1.7  & 84.3 $\pm$ 2.7   \\ 
    halfcheetah-medium-expert-v2 
    & 85.4 $\pm$ 3.3  & 19.2 $\pm$ 4.1  & 82.2 $\pm$ 5.2  & \colorbox{myblue}{94.5} $\pm$ 0.6  & 83.4 $\pm$ 5.3   \\ 
    hopper-medium-expert-v2 
    & 88.7 $\pm$ 2.9  & 14.5 $\pm$ 12.5  & 97.0 $\pm$ 7.6  & \colorbox{myblue}{102.5} $\pm$ 4.5  & 85.9 $\pm$ 6.6   \\ 
    walker2d-medium-expert-v2 
    & 110.9 $\pm$ 0.6  & 1.0 $\pm$ 2.5  & 110.1 $\pm$ 0.8  & 110.8 $\pm$ 0.4  & \colorbox{myblue}{111.1} $\pm$ 0.6   \\ 
    \hline
    \textbf{Gym Average} 
    & \textbf{56.4} $\pm$ 2.2  & \textbf{11.6} $\pm$ 6.4  & \textbf{57.2} $\pm$ 3.9
    & \textbf{63.2} $\pm$ 2.9  & \colorbox{myblue}{\textbf{64.9}} $\pm$ 3.3 
    \\ 
    \hline
    antmaze-umaze-v0 
    & \colorbox{myblue}{94.4}  $\pm$ 2.7  & 0  & 0  & 72.8  $\pm$ 36.8  & 90.4  $\pm$ 5.2   \\
    antmaze-umaze-diverse-v0
    & 51.0  $\pm$ 16.8  & 0  & 0  & 62.5  $\pm$ 31.2  & \colorbox{myblue}{75.0}  $\pm$ 19.0   \\
    antmaze-medium-play-v0
    & 1.4  $\pm$0.8  & 0   & 0   & 0  & \colorbox{myblue}{62.0}  $\pm$ 10   \\
    antmaze-medium-diverse-v0
    & 1.0  $\pm$ 1.9  & 0   & 0   & 0.3  $\pm$ 0.4  & \colorbox{myblue}{54.5}  $\pm$ 23.3   \\
    antmaze-large-play-v0
    & 0    & 0    & 0    & 0   & 0     \\ 
    antmaze-large-diverse-v0
    & 0    & 0    & 0    & 0    & \colorbox{myblue}{9.4}  $\pm$ 8.4   \\ 
    \hline
    \textbf{AntMaze Average}
    & \textbf{24.6}  $\pm$ 3.7  & \textbf{0}   & \textbf{0}   & \textbf{22.6}  $\pm$ 11.4  & \colorbox{myblue}{\textbf{48.6}}  $\pm$ 11.0   \\ 
    \hline
    pen-human-v1 
    & 84.8 $\pm$ 11.2  & 4.3 $\pm$ 7.4  & 8.5 $\pm$ 10.1
    & 79.1 $\pm$ 14.5  & \colorbox{myblue}{101.2} $\pm$ 11.5 
    \\ 
    pen-cloned-v1 
    & 56.2 $\pm$ 16.3  & 4.3 $\pm$ 7.4  & 57.2 $\pm$ 29.5
    & 66.8 $\pm$ 11.1  & \colorbox{myblue}{73.5} $\pm$ 13.0 
    \\ 
    \hline
    \textbf{Adroit Average} 
    & \textbf{70.5} $\pm$ 13.7  & \textbf{4.3} $\pm$ 7.4  & \textbf{32.8} $\pm$ 19.8
    & \textbf{72.9} $\pm$ 12.8  & \colorbox{myblue}{\textbf{87.4}} $\pm$ 12.2 
    \\ 
    \hline
    \hline
    \textbf{Average} 
    & \textbf{48.3} $\pm$ 3.8  & \textbf{7.4} $\pm$ 4.6  & \textbf{37.6} $\pm$ 4.3
    & \textbf{52.0} $\pm$ 6.4  & \colorbox{myblue}{\textbf{62.2}} $\pm$ 6.5 
    \\
    \specialrule{0.12em}{0pt}{0pt}
  \end{tabular}
  \label{table: results_on_TD3+BC}
  \vspace{-5pt}
\end{table*}

\begin{figure*}
\centering
    \includegraphics[width=0.95\textwidth]{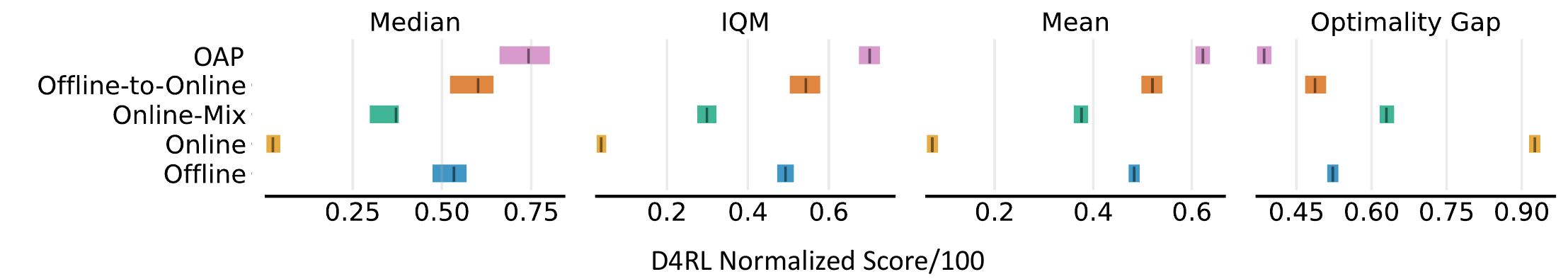}
    \vskip -0.06 in
    \caption{
    Statistical results of different training schemes instantiated on TD3+BC by rliable~\cite{rliable} over 5 random seeds.
    }
    \label{fig: rliable}
\end{figure*}

\begin{figure*}
\centering
    \includegraphics[width=0.85\textwidth]{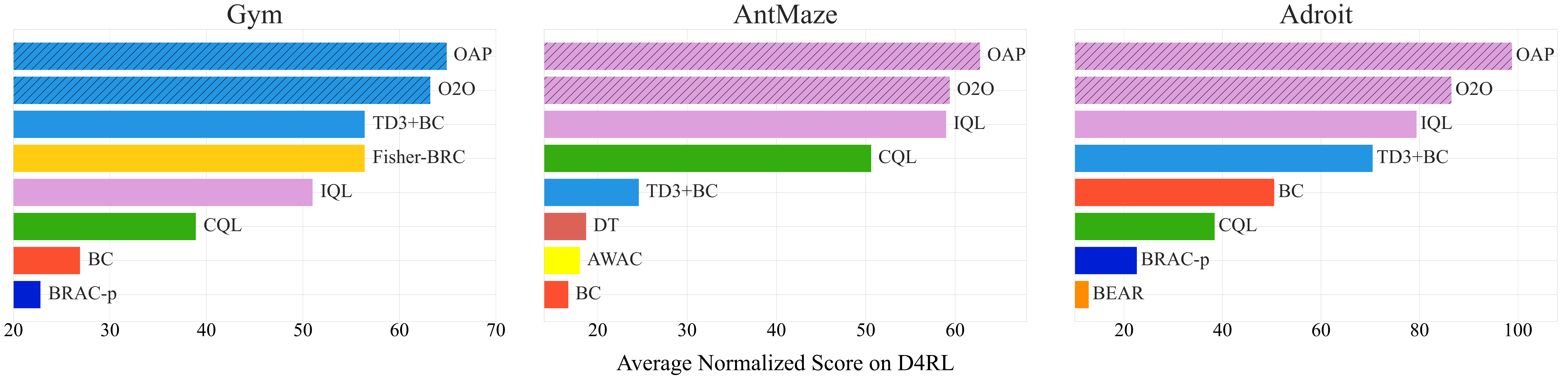}
    \vskip -0.06 in
    \caption{Results of the Offline-to-Online (O2O) and Offline-with-Action-Preferences (OAP) schemes instantiated on SOTA offline RL algorithms. OAP further improves the best-performing algorithms in all three domains. We reproduce IQL and TD3+BC following author-provided implementations, and other offline results are from \cite{td3+bc,IQL}.}
    \label{fig: sota}
\end{figure*}

We investigate different training schemes on various domains to find a better way of utilizing real-world resources. 
Firstly, a range of dataset compositions and training schemes are introduced in \cref{sec: experiments_setup}. 
Then, we instantiate these schemes on a popular offline RL algorithm and compare their performances in \cref{sec: experiments_Comparisons
among Schemes}. 
Finally, the high-performing schemes are instantiated on more state-of-the-art algorithms for generalized conclusions in \cref{sec: experiments_Results on SOTA Baselines}.   

\subsection{Setup}
\label{sec: experiments_setup}
\textbf{Datasets.} We consider three different domains of tasks in D4RL~\cite{Fu2020D4RLDF} benchmark: Gym, AntMaze, and Adroit. 
The Gym-Mujoco tasks include datasets in various environments (\textit{e.g.}, halfcheetah, hopper, and walker) with different qualities (\textit{e.g.}, random, medium, medium-replay, and medium-expert). 
AntMaze and Adroit tasks are more challenging, and even online RL algorithms struggle to complete them.
The AntMaze domain involves navigation tasks that require an 8-DoF ``Ant'' quadruped robot to reach a goal location. 
There are three maze layouts (umaze, medium, large) with different location types (play and diverse).
The Adroit datasets are mostly collected by human behavior and aim at controlling a 24-DoF robotic hand.

\textbf{Schemes.} Given that an offline dataset and limited interactions/queries are available, there are five schemes of training an agent, as shown in \cref{table: setting_comparison}.
The \textit{Offline} scheme requires pre-collected offline data, and the agent only learns from that fixed dataset.
The \textit{Online} scheme trains the agent through real-time interactions with the environment, where the state transition function and reward function are necessary.
The \textit{Online-Mix} scheme is similar to the \textit{Online} one except for adding the offline data into the online replay buffer.
The \textit{Offline-to-Online (O2O)} scheme pre-trains the agent in an \textit{Offline} way and fine-tunes it in an \textit{Online-Mix} way. 
The \textit{Offline-with-Action-Preferences (OAP)} scheme learns a policy from the offline dataset and periodically queries a blackbox model for action preferences.
For all the schemes above, we consider online interaction steps or action preference queries limited to 100k, which is usually accessible and acceptable in real-world applications.
More implementation details are in \cref{appendix: Implementation Details}.


\subsection{Comparisons among Schemes}
\label{sec: experiments_Comparisons among Schemes}

We investigate the different schemes presented above using the popular offline RL algorithm, TD3+BC~\cite{td3+bc}. 
Comprehensive experiments are conducted to evaluate each scheme on various tasks, as shown in \cref{table: results_on_TD3+BC}.
Firstly, we observe that the Online scheme fails on all tasks because of insufficient data (100k).
Secondly, the performances of the Online-Mix scheme are comparable to that of the Offline scheme on Gym tasks but show a tremendous drop in more challenging domains (AntMaze and Adroit).
As shown in previous work~\cite{fu2019diagnosing,kumar2020discor,yue2022boosting,anonymous2023actorcritic}, the distributional gap between offline data and newly-added online data may harm the training.
Thirdly, the Offline-to-Online scheme improves the pre-trained policy on some tasks but still suffers from the distributional gap on some challenging tasks (\textit{e.g.} antmaze-umaze and pen-human).
Finally, compared to other schemes, our method OAP dramatically improves upon the Offline baseline on all three domains.
In addition, compared to previous work based on the O2O scheme, our method has the following advantages: $(1)$ it does not require real-world interactions nor $(2)$ a well-designed reward function.

\textbf{Statistical validation.}
In addition to the point estimates of aggregate performance in \cref{table: results_on_TD3+BC}, we present a more rigorous evaluation in \cref{fig: rliable}. 
These metrics from rliable~\cite{rliable} increase the results' confidence by accounting for the statistical uncertainty in a handful of runs.
Four metrics are considered: median, interquartile mean (IQM), mean, and optimality gap.
IQM (also called 25\% trimmed mean) and optimality gap are robust alternatives to median and mean respectively.
Higher mean, median and IQM scores and lower optimality gap are better.
Results in \cref{fig: rliable} statistically support the conclusions in \cref{table: results_on_TD3+BC} and validate the superiority of our OAP method.

\subsection{Results on SOTA Baselines}
\label{sec: experiments_Results on SOTA Baselines}
To validate the generalization of our method, we report results on various D4RL domains\cite{Fu2020D4RLDF} using state-of-the-art algorithms with Offline-to-Online and OAP schemes.
For policy regularization-based methods, we consider BC, BEAR\cite{BEAR}, BRAC\cite{BRAC}, AWAC\cite{AWAC}, Fisher-BRC\cite{Fisher-BRC}, TD3+BC\cite{td3+bc}, and IQL\cite{IQL}.
For Q-value constraint and sequence modeling methods, we include CQL~\cite{CQL} and Decision Transformer (DT)~\cite{DT}.

We first compare different offline RL algorithms on the Offline scheme and select the one with the highest score.
Then, the most powerful algorithm in this domain is equipped with the Offline-to-Online (O2O) or Offline-with-Action-Preferences (OAP) scheme. 
As shown in \cref{fig: sota}, the SOTA algorithm on Gym is TD3+BC, and IQL is the best for AntMaze and Adroit.
We can observe that the two schemes both improve the performance of the SOTA algorithm.
Meanwhile, our OAP scheme brings stable performance gain compared to the O2O scheme on all three domains, especially on the harder AntMaze and Adroit tasks.

\section{Discussion}
This section presents in-depth investigations of our proposed method.
Experiments are based on TD3+BC \cite{td3+bc} and the benchmark is D4RL~\cite{Fu2020D4RLDF}.
In particular, we investigate OAP's four main components: the blackbox policy, the adjusted policy constraint, the periodical queries, and the RankNet. 
Hence, analytical experiments are designed around the following four~questions.

\subsection{Can OAP work with faulty annotations?}

\begin{figure}[t]
\centering
\includegraphics[width=0.47\textwidth]{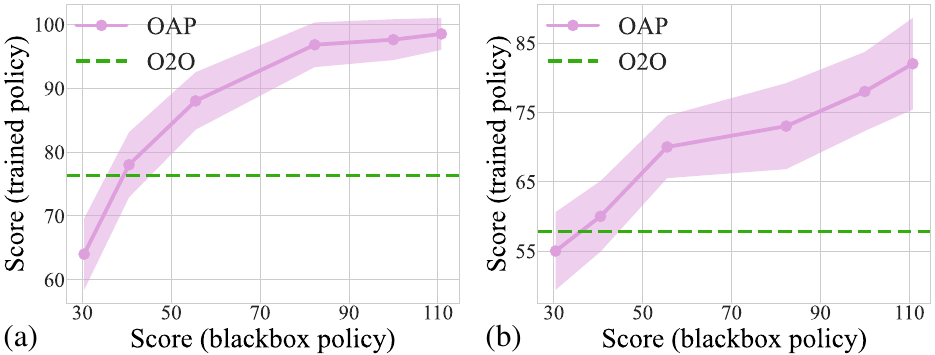}
    \caption{OAP queries a blackbox policy ($\pi^*$) for action preferences. Blackbox policies of various quality are compared in this figure (a. hopper-medium-replay, b. hopper-medium). Results validate that even sub-optimal $\pi^*$ can lead to significant performance improvement, which makes OAP more practical in the real world.
    }
    \label{fig: ablation_expert}
\end{figure}

We denote the policy that provides action preferences as the \textit{blackbox policy} ($\pi^*$) and the one that is trained with OAP as the \textit{trained policy} ($\pi$).
The policy that achieves a full normalized D4RL score ($\ge$ 100) can be considered expert quality.
In real-world applications, the proprietary preference model usually performs at the expert quality, and it's feasible to adopt it as $\pi^*$.
Meanwhile, considering faulty annotations and how OAP works correspondingly are equally valuable since security and stability are always highly prioritized in practice.
As shown in \cref{fig: ablation_expert}, policies of different quality (\textit{i.e.}, high or low D4RL score) are adopted as $\pi^*$ in, OAP and the performance of trained policies are compared.
Taking the gym tasks where OAP works most effectively as examples, we can observe that OAP brings more considerable performance improvement to the trained policy than O2O even with poor-performing balckbox policy (score $<$ 50).
These results suggest that OAP has a high tolerance for faulty annotations and works well with sub-optimal $\pi^*$, which coincides with our theoretical analyses in \cref{sec: method Theoretical Analysis}.

\begin{figure*}[ht]
\centering
\includegraphics[width=0.91\textwidth]{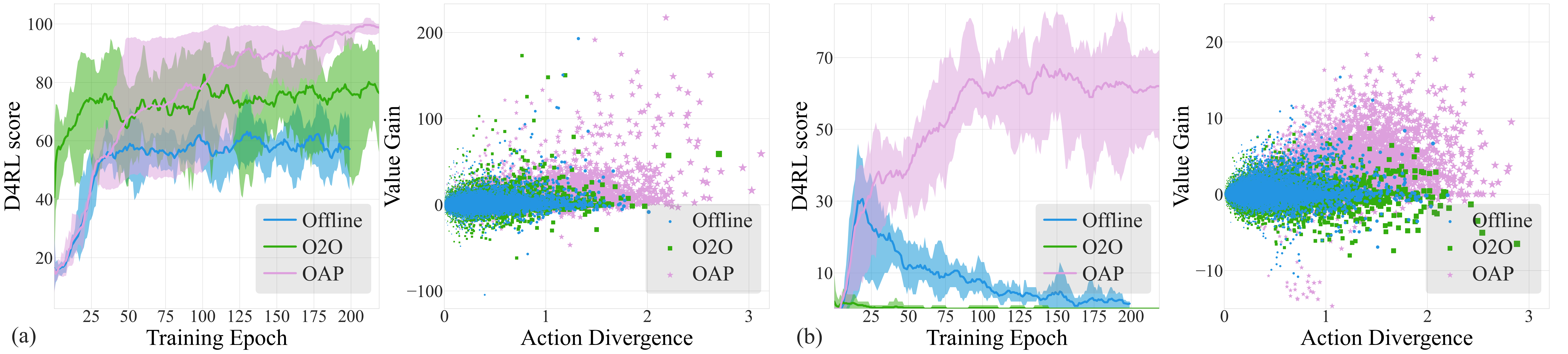}
    \vskip -0.1 in
    \caption{The training curve (Training Epoch \textit{vs.} D4RL Score) and policy constraint quality (Action Divergence \textit{vs.} Value Gain) of two typical tasks in Gym (a. hopper-medium-replay) and AntMaze (b. antmaze-medium-play) domains. OAP performs better than O2O because its out-of-distribution data (\textit{i.e.,} data with large action divergences) usually have positive and high value gains.}
    \label{fig: ablationQ}
    \vspace{-5pt}
\end{figure*}

\subsection{Do action preferences help policy constraint?}
\cref{sec: method Adjusted Policy Constraint} shows that OAP can adaptively constrain the policy on either actions given by the dataset or by the learned policy.
In particular, given a sample $(s_t,a_t)$ from the offline dataset, the preference model selects between $a_t$ and the learned action $\pi(s_t)$. 
If $a_t$ is preferred, we constrain the policy on the offline dataset. Conversely, if $\pi(s_t)$ is preferred, the constraint is loosened to the trained policy's vicinity.

We adopt \textit{action divergence} and \textit{value gain} to measure the quality of policy constraint.
For a well-trained policy $\pi$ and a pair $(s_t,a_t)$ in the offline dataset, the action divergence calculates the distance of $\pi(s_t)$ and $a_t$ for the whole dataset.
Given an accurate value function $Q^*(s,a)$, the value gain is $Q^*(s_t,\pi(s_t))-Q^*(s_t,a_t)$, representing how many values the policy gains by conducting $\pi(s_t)$ instead of $a_t$.
A large action divergence shows that the policy takes an aggressive move, and a large value gain means this move is worthy.

In \cref{fig: ablationQ}, the Offline, O2O, and OAP schemes are compared, and there are three observations.
Firstly, the policy trained from OAP has the largest action divergence, and that from the Offline scheme has the smallest, which means OAP learns more OOD actions.
Secondly, compared to the Offline scheme, O2O learns a more aggressive policy (\textit{i.e.}, more divergent actions) but meanwhile suffers from the overestimation problem (\textit{i.e.}, more negative value gains).
Thirdly, the most divergent actions in OAP usually have large positive value gains, which means the policy is beneficially aggressive.
Hence, it is validated that OAP facilitates more adaptive policy constraints by encouraging high-value divergences and restraining harmful ones.

\subsection{What if queries were focused instead of spaced?}
O2O allows interactions with the environment after training on the offline dataset, which is a typical procedure of pretrain-finetune.
Contrastively, OAP spaces the queries throughout the training process because we assume that timely corrections of policy constraints are better than changing a convergent policy.
Hence, we investigate whether spacing the queries matters or not and whether O2O also benefits from spaced interactions or not.

\begin{figure}[ht]

\centering
    \includegraphics[width=0.46\textwidth]{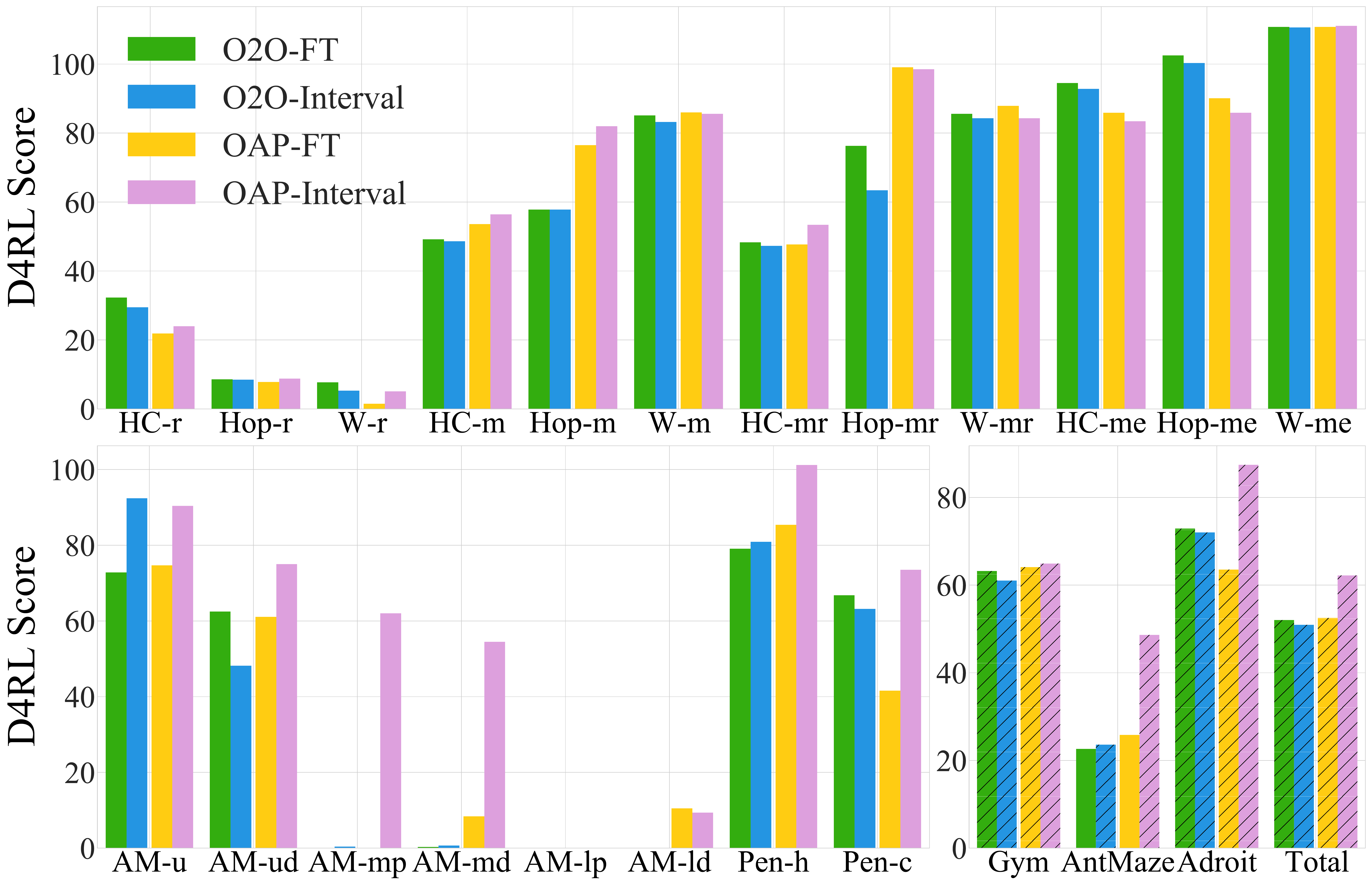}
    \vskip -0.1 in
    \caption{Normalized D4RL score of O2O, OAP, and variants on Gym, AntMaze, and Adroit tasks.
    \protect\footnotemark[3]{} 
    \textit{FT} means pre-training on offline data and fine-tuning with interactions/queries. \textit{Interval} means spreading interactions/queries throughout the training process.}
    \label{fig: ablation_interval}
\end{figure}

\footnotetext[3]{HC = Halfcheetah, Hop = Hopper, W = Walker, r = random, m = medium, mr = medium-replay, me = medium-expert; AM = antmaze, u = umaze, m = medium, l = large, d = diverse, p = play, h = human, c = cloned.}

We denote the pretrain-finetune procedure as \textit{FT} and the periodical interactions/queries as \textit{Interval}.
In \cref{fig: ablation_interval}, \textit{O2O} means interacting with the environment, and \textit{OAP} means querying for action preferences.
O2O-FT and OAP-Interval correspond to the original O2O and OAP.
It is noted that all four methods in \cref{fig: ablation_interval}  conform to the 100k limitations of interactions/queries. 
Results show that O2O-FT and O2O-Interval generally have comparable performances in three domains.
OAP-Interval performs similarly to OAP-FT on Gym but is much better on harder AntMaze and Adroit.
Therefore, we suggest that spaced queries are~necessary for OAP, but periodical interactions cannot benefit O2O.

\subsection{Is there a necessity for RankNet?}
During training, since the 100k queries only cover a small set of samples, it is natural to assume that its effect on the policy constraint is limited. 
As described in Section \ref{sec: method Pseudo Query with RankNet}, RankNet extends the number of queries by pseudo-annotating the unqueried samples.
Ideally, we would like RankNet to perfectly learn to rank two actions and provide expert preferences. 
Specifically, we compare two variants of OAP to validate the effectiveness of RankNet.
OAP(inf) allows action preference queries for each sample in the offline dataset instead of the 100k limitation in the normal OAP.
OAP(w/RN) maintains the 100k limitation of queries, and the unqueried samples cannot have pseudo annotations from RankNet.
As shown in \cref{table: ablation_query_samples}, OAP(inf) performs slightly better than OAP, while the performance of OAP(w/RN) has a significant drop.
The former result suggests that RankNet successfully learns the pattern of an expert's action preferences in most cases.   
The latter result shows that the 100k queries are not enough, and RankNet is necessary for OAP.

\begin{table}[t]
  \small
  \caption{Average normalized D4RL score~\cite{Fu2020D4RLDF} over the final 10 evaluations and 5 random seeds. OAP(inf) means OAP with unlimited times of queries. OAP(w/RN) means OAP without pseudo queries from RankNet but with limited querying times. 
  }
  \centering
  \begin{tabular}{lccc}
    \specialrule{0.12em}{0pt}{0pt}
	Dataset
	& OAP 
	& OAP (inf)
	& OAP (w/ RN) 
	\\ \hline
    HC-r
    & 24.0 $\pm$ 1.6  
    & 22.0 $\pm$ 1.4 
    & 11.0  $\pm$ 1.1 
    \\
    Hop-r
    & 8.8 $\pm$ 1.8 
    & 13.2  $\pm$ 7.4 
    & 8.1  $\pm$ 0.8 
    \\
    W-r
    & 5.1 $\pm$ 5.1 
    & 2.3  $\pm$ 2.0 
    & 1.9  $\pm$ 0.9 
    \\
    HC-m
    & 56.4 $\pm$ 4.3  
    & 59.2  $\pm$ 1.5 
    & 48.2  $\pm$ 0.3 
    \\ 
    Hop-m
    & 82.0 $\pm$ 6.6
    & 92.8  $\pm$ 4.1 
    & 45.9  $\pm$ 1.5 
    \\ 
    W-m
    & 85.6 $\pm$ 1.2   
    & 86.6  $\pm$ 0.3 
    & 84.8  $\pm$ 2.4 
    \\ 
    HC-mr
    & 53.4 $\pm$ 1.9
    & 51.3  $\pm$ 0.6 
    & 28.4  $\pm$ 3.1  
    \\ 
    Hop-mr
    & 98.5 $\pm$ 2.5
    & 101.9  $\pm$ 2.0 
    & 31.6  $\pm$ 3.5 
    \\ 
    W-mr
    & 84.3 $\pm$ 2.7
    & 84.9  $\pm$ 9.9 
    & 74.8  $\pm$ 5.5 
    \\ 
    HC-me
    & 83.4 $\pm$ 5.3
    & 84.1  $\pm$ 3.2 
    & 84.3  $\pm$ 5.3 
    \\ 
    Hop-me
    & 85.9 $\pm$ 6.6
    & 92.2  $\pm$ 8.2 
    & 80.6  $\pm$ 2.6 
    \\ 
    W-me
    & 111.1 $\pm$ 0.6
    & 109.4  $\pm$ 1.5 
    & 111.0  $\pm$ 0.4 
    \\ 
    \hline
    \textbf{Avg.} 
    & \textbf{64.9} $\pm$ 3.3 
    & \textbf{66.7}  $\pm$ 3.5 
    & \textbf{50.9}  $\pm$ 2.3 
    \\
    \specialrule{0.12em}{0pt}{0pt}
  \end{tabular}
  \label{table: ablation_query_samples}
  \vspace{-10pt}
\end{table}

\section{Related Work}
\textbf{Offline RL.} As aforementioned, offline RL suffers from \textit{distributional shift} caused by the gap between pre-training and OOD data.
Existing methods generally constrain or regularize the learned policy to limit the deviation from the behavior policy.
This may be implemented by an explicit density model~\cite{BRAC,Off_Policy_Deep_Reinforcement_Learning_without_Exploration,BEAR,EMaQ}, implicit divergence constraints~\cite{reward_weighted_regression,Advantage_Weighted_Regression,AWAC,Critic_Regularized_Regression,IQL}, pessimistic estimations of state-action values~\cite{CQL,Fisher-BRC}, or directly adding a behavior cloning term to the policy improvement loss~\cite{DDPG_BC,PPO+BC,td3+bc}.
In the same way that powerful computer vision and NLP models are often pre-trained on large, general-purpose datasets and then fine-tuned on task-specific data, practical instantiations of RL for real-world applications usually involve pre-training on the offline dataset and then fine-tuning with online interactions~\cite{AWAC,IQL}.

\textbf{Preference Learning.} 
Roughly speaking, preference learning involves inducing predictive preference models from empirical data~\cite{preference_survey}.
A preference learning task consists of some set of items for which
preferences are known, and the task is to learn a function that predicts preferences for a new set of items, or the same set of items in a different context~\cite{Preference_Learning}.
One of the most extensively studied tasks in preference learning is learning to rank (LTR).
The commonly used LTR algorithms are mainly pointwise~\cite{mcrank}, pairwise~\cite{ranknet}, and listwise~\cite{lambdarank,lambdamart}.
The pointwise and pairwise approaches treat relevance degree as real values or categories, while the pairwise methods reduce ranking to classifying the order between each pair.

In preference-based RL (PbRL), the main problem is learning a policy using preferences between states, actions, or trajectories without any numeric reward signal.
It replaces reward values in online RL by preferences to better elicit human opinions on the target objective, especially when numerical reward values are hard to design or interpret.
The process usually involves two actors: an agent that acts according
to a given policy and an expert evaluating its behavior~\cite{preference_survey}, which is similar to OAP.
On the other hand, unlike PbRL algorithms, our method only uses a small amount of preference information and learns from a static offline dataset.
Preferences are used to~help agents distinguish good or bad policy improvement rather than directly taken as training feedback or reformulated as reward functions.
As far as we know, OAP takes the first step to utilize preference learning for boosting offline RL and achieving more extensive success than online fine-tuning~methods.

\section{Conclusion}
\label{sec: conclusion}
We present Offline-with-Action-Preferences (OAP), a general training scheme for practical offline RL that brings higher returns than online fine-tuning but dispenses with the risk of real-world interactions.
To our knowledge, OAP is the first method that avoids the challenges of online fine-tuning and meanwhile achieves better performance than previous methods that leverage online fine-tuning.
This has a number of significant benefits since online fine-tuning is usually essential for offline agents in real-world applications.
Firstly, our method is more practical for high-stake scenarios like healthcare and self-driving because interactions with the environment are not required.
Secondly, we don't require well-defined reward signals as online feedback, which can be hard to design in many tasks.  
Thirdly, OAP exploits limited queries efficiently to ensure it is economical and accessible.
Finally, besides the safety and efficiency of this method, we show that it attains superior performance across D4RL tasks, either compared to other high-risk training schemes or instantiated on more state-of-the-art algorithms.
This work is a step towards our larger vision of more practical RL, where the key is to address the uncertainty and risk of online interactions.
We hope it can inspire future research, such as offline RL with weak action preferences, intricate policy constraints, state/trajectory preferences, revised Q-learning, and adaptive uncertainty estimation.

\section*{Acknowledgments}
This work is supported in part by the National Key R\&D Program of China under Grant 2022ZD0114900, the National Natural Science Foundation of China under Grants 62022048 and 62276150, and the Guoqiang Institute of Tsinghua University. We also appreciate the generous donation of computing resources by High-Flyer AI.

\bibliography{ref}
\bibliographystyle{icml2023}

\newpage
\appendix
\onecolumn
\section{Theoretical Proofs}
\label{appendix: Theoretical Proofs}
\subsection{Proof of \cref{prop: perfect case}}~\label{appendix: the proof of the perfect case}
We first start with a lemma considering the policy improvement as follows:
\begin{lemma}~\label{lemma: policy improvement}
Given any two policy $\pi_1$ and $\pi_2$,
\begin{equation}
    \eta(\pi_1) - \eta(\pi_2) = \int_{s\in\mathcal{S}} \rho_{\pi_1}(s) ( Q_{\pi_2}(s, \pi_1(s)) - V_{\pi_2}(s)) \diff s
\end{equation}
\end{lemma}
\begin{proof}
The derivation of this lemma is related to \citet{TRPO} and \citet{kakade2002approximately}.

According to \citet{kakade2002approximately}, 
\begin{equation}
\eta(\pi_1)=\eta(\pi_2)+\mathbb{E}_{\tau \sim \pi_1}\left[\sum_{t=0}^{\infty} \gamma^t Q_{\pi_2}\left(s_t, a_t\right) - V_{\pi_2}\left(s_t\right)\right].
\end{equation}
Here, $\mathbb{E}_{\tau \sim \pi_1}$ denotes sampling trajectories with the policy $\pi_1$.

It follows that 
\begin{align}
\eta(\pi_1) & =\eta(\pi_2)+\sum_{t=0}^{\infty} \int_{s\in \mathcal{S}} P\left(s_t=s | \pi_1\right) \gamma^t (Q_{\pi_2}(s, \pi_1(s)) - V_{\pi_2}(s)) \diff s \\
& =\eta(\pi_2)+\int_{s\in\mathcal{S}} \sum_{t=0}^{\infty} \gamma^t P\left(s_t=s | \pi_1\right) (Q_{\pi_2}(s, \pi_1(s)) - V_{\pi_2}(s)) \diff s\\
& =\eta(\pi_2)+\int_{s\in\mathcal{S}} \rho_{\pi_1}(s) (Q_{\pi_2}(s, \pi_1(s)) - V_{\pi_2}(s)) \diff s.
\end{align}
Therefore \cref{lemma: policy improvement} is proven.
\end{proof}

According to \cref{lemma: policy improvement}, it follows that 
\begin{align}
    &\eta(\Tilde{\pi}_\beta) - \eta(\pi^*) = \int_{s\in\mathcal{S}} \rho_{\Tilde{\pi}_\beta}(s) ( Q^*(s, \Tilde{\pi}_\beta(s)) - V^*(s)) \diff s \label{eq: policy improvement 1} \\
    &\eta(\pi_\beta) - \eta(\pi^*) =  \int_{s\in\mathcal{S}} \rho_{\pi_\beta}(s) ( Q^*(s, \pi_\beta(s)) - V^*(s)) \diff s \label{eq: policy improvement 2}
\end{align}
Combining \cref{eq: policy improvement 1} and \cref{eq: policy improvement 2}, we can infer that $\eta(\Tilde{\pi}_\beta) - \eta(\pi_\beta)$ satisfies:
\begin{align}
    \eta(\Tilde{\pi}_\beta) - \eta(\pi_\beta) &= (\eta(\Tilde{\pi}_\beta) - \eta(\pi^*)) - (\eta(\pi_\beta) - \eta(\pi^*))\\
    & = \int_{s\in\mathcal{S}} \rho_{\Tilde{\pi}_\beta}(s) ( Q^*(s, \Tilde{\pi}_\beta(s)) - V^*(s)) \diff s
    -
    \int_{s\in\mathcal{S}} \rho_{\pi_\beta}(s) ( Q^*(s, \pi_\beta(s)) - V^*(s)) \diff s \label{eq: before approximation in the perfect case} \\
    & \approx  \int_{s\in\mathcal{S}} \rho_{\pi_\beta}(s)( Q^*(s, \Tilde{\pi}_\beta(s)) - Q^*(s, \pi_\beta(s)) ) \diff s \label{eq: approximation in the perfect case}
\end{align}
The derivation from \cref{eq: before approximation in the perfect case} to \cref{eq: approximation in the perfect case} is because there are only a few queries compared with the huge number of offline data, it follows $\rho_{\Tilde{\pi}_\beta}(s) \approx \rho_{\pi_\beta}(s)$.

Based on \cref{eq: pr_true}, it holds that $\forall s \in \mathcal{S}, Q^*(s, \Tilde{\pi}_\beta(s)) \ge Q^*(s, \pi_\beta(s))$. 
Considering $\forall s \in \mathcal{S}, \rho_{\pi_\beta}(s) \ge 0$, it follows that 
\begin{equation}
    \eta(\Tilde{\pi}_\beta) - \eta(\pi_\beta) \approx  \int_{s\in\mathcal{S}} \rho_{\pi_\beta}(s)( Q^*(s, \Tilde{\pi}_\beta(s)) - Q^*(s, \pi_\beta(s)) ) \diff s \ge 0.
\end{equation}
Noting that $\int_{s\in\mathcal{S}} \rho_{\pi_\beta}(s) \cdot \diff s$ is equivalent to $\mathbb{E}_{s\sim \mathcal{D}}[\cdot]$, \cref{prop: perfect case} follows. \textbf{Q.E.D.}

\subsection{Proof of \cref{prop: imperfect case}}~\label{appendix: the proof of the imperfect case}
For the imperfect query case, we denote that 
\begin{equation}
   Q^*(s, a) = \hat{Q}^*(s, a) + \delta(s, a),  \label{eq: approximation error}
\end{equation}
where $\delta(s, a)$ is the approximation error of the estimated state-value function $\hat{Q}^*(s, a)$.

Combining  \cref{eq: approximation in the perfect case} and \cref{eq: approximation error}, it follows that 
\begin{align}
    \eta(\Tilde{\pi}_\beta) - \eta(\pi_\beta) & \approx  \int_{s\in\mathcal{S}} \rho_{\pi_\beta}(s)( Q^*(s, \Tilde{\pi}_\beta(s)) - Q^*(s, \pi_\beta(s)) ) \diff s \\
    &=  \int_{s\in\mathcal{S}} \rho_{\pi_\beta}(s)\left[  (\hat{Q}^*(s, \Tilde{\pi}_\beta(s)) + \delta(s, \Tilde{\pi}_\beta(s))) - (\hat{Q}^*(s, \pi_\beta(s)) + \delta(s, \pi_\beta(s))) \right] \diff s \\
    & = \int_{s\in\mathcal{S}} \rho_{\pi_\beta}(s) \left[\hat{Q}^*(s, \Tilde{\pi}_\beta(s)) - \hat{Q}^*(s, \pi_\beta(s))\right] \diff s - 
\int_{s\in\mathcal{S}} \rho_{\pi_\beta}(s)[\delta(s, \Tilde{\pi}_\beta(s))- \delta(s, \pi_\beta(s))] \diff s. \label{eq: Q difference and delta difference}
\end{align}
Considering the condition in \cref{prop: imperfect case} that $D_{\mathrm{TV}}^{\Tilde{\pi}_\beta}(\Hat{Q}^*, Q^*) \le \Tilde{\alpha}, D_{\mathrm{TV}}^{\pi_\beta}(\Hat{Q}^*, Q^*) \le \alpha$, we have 
\begin{align}
   &\forall s\in \mathcal{S}, \quad |\delta(s, \Tilde{\pi}_\beta(s))| = \left|Q^*(s, \Tilde{\pi}_\beta(s)) - \hat{Q}^*(s, \Tilde{\pi}_\beta(s))\right| \le D_{\mathrm{TV}}^{\Tilde{\pi}_\beta}(\Hat{Q}^*, Q^*) \le \Tilde{\alpha}, \label{eq: alpha-tilde}\\
   &\forall s\in \mathcal{S}, \quad |\delta(s, \pi_\beta(s))| = \left|Q^*(s, \pi_\beta(s)) - \hat{Q}^*(s, \pi_\beta(s))\right| \le D_{\mathrm{TV}}^{\pi_\beta}(\Hat{Q}^*, Q^*) \le \alpha. \label{eq: alpha}
\end{align}
Therefore,
\begin{equation}
    \forall s\in \mathcal{S}, \quad |\delta(s, \Tilde{\pi}_\beta(s)) - \delta(s, \pi_\beta(s))| \le |\delta(s, \Tilde{\pi}_\beta(s))| + |\delta(s, \pi_\beta(s))| = \Tilde{\alpha} + \alpha.  \label{eq: upper bound of delta difference}
\end{equation}

To prepare for further derivation, We need to introduce \cref{lemma: D-TV}.
\begin{lemma}[\citet{wilmer2009markov}] \label{lemma: D-TV}
    Suppose that $\mu$ and $\nu$ are two probability distributions on $\mathcal{X}$, then
    \begin{equation}
        \max_{x \in \mathcal{X}} | \mu(x) - \nu(x) | = \frac{1}{2} \sum_{x\in \mathcal{X}} | \mu(x) - \nu(x) |. 
    \end{equation}
\end{lemma}
\begin{proof}
    Please refer to Proposition~4.2 in  \citet{wilmer2009markov}.
\end{proof}

Denote $\overline{\rho}_{\pi_\beta} = \sup \{\rho_{\pi_\beta}(s), s\in\mathcal{S}\}$.
Based on \cref{eq: Q difference and delta difference},  it follows 
\begin{align}
    \eta(\Tilde{\pi}_\beta) - \eta(\pi_\beta) & \approx \int_{s\in\mathcal{S}} \rho_{\pi_\beta}(s) \left[\hat{Q}^*(s, \Tilde{\pi}_\beta(s)) - \hat{Q}^*(s, \pi_\beta(s))\right] \diff s - 
\int_{s\in\mathcal{S}} \rho_{\pi_\beta}(s)[\delta(s, \Tilde{\pi}_\beta(s))- \delta(s, \pi_\beta(s))]\diff s\\
& =  \mathbb{E}_{s\sim \mathcal{D}} \left[\hat{Q}^*(s, \Tilde{\pi}_\beta(s)) - \hat{Q}^*(s, \pi_\beta(s))\right]
- 
\overline{\rho}_{\pi_\beta} \int_{s\in\mathcal{S}} \frac{\rho_{\pi_\beta}(s)}{\overline{\rho}_{\pi_\beta}}[\delta(s, \Tilde{\pi}_\beta(s))- \delta(s, \pi_\beta(s))]\diff s \\
& \ge \mathbb{E}_{s\sim \mathcal{D}} \left[\hat{Q}^*(s, \Tilde{\pi}_\beta(s)) - \hat{Q}^*(s, \pi_\beta(s))\right]
- 
\overline{\rho}_{\pi_\beta} \int_{s\in\mathcal{S}} \left|\frac{\rho_{\pi_\beta}(s)}{\overline{\rho}_{\pi_\beta}}\left[\delta(s, \Tilde{\pi}_\beta(s))- \delta(s, \pi_\beta(s))\right]\right| \diff s.
\end{align}
Due to $\forall s\in\mathcal{S}, \frac{\rho_{\pi_\beta}(s)}{\overline{\rho}_{\pi_\beta}} \in [0, 1]$, 
\begin{align}
    \eta(\Tilde{\pi}_\beta) - \eta(\pi_\beta) & \gtrsim \mathbb{E}_{s\sim \mathcal{D}} \left[\hat{Q}^*(s, \Tilde{\pi}_\beta(s)) - \hat{Q}^*(s, \pi_\beta(s))\right]
- 
\overline{\rho}_{\pi_\beta} \int_{s\in\mathcal{S}} \left|\frac{\rho_{\pi_\beta}(s)}{\overline{\rho}_{\pi_\beta}}[\delta(s, \Tilde{\pi}_\beta(s)) - \delta(s, \pi_\beta(s))]\right| \diff s \\
& \ge \mathbb{E}_{s\sim \mathcal{D}} \left[\hat{Q}^*(s, \Tilde{\pi}_\beta(s)) - \hat{Q}^*(s, \pi_\beta(s))\right]
-
\overline{\rho}_{\pi_\beta} \int_{s\in\mathcal{S}} \left|\delta(s, \Tilde{\pi}_\beta(s))- \delta(s, \pi_\beta(s))\right| \diff s \label{eq: before lamma D-TV} \\
& = \mathbb{E}_{s\sim \mathcal{D}} \left[\hat{Q}^*(s, \Tilde{\pi}_\beta(s)) - \hat{Q}^*(s, \pi_\beta(s))\right]
-
2 \overline{\rho}_{\pi_\beta} \max_{s\in\mathcal{S}} \left|\delta(s, \Tilde{\pi}_\beta(s))- \delta(s, \pi_\beta(s))\right| \label{eq: after lamma D-TV} \quad \text{(\cref{lemma: D-TV})} \\
& \ge \mathbb{E}_{s\sim \mathcal{D}} \left[\hat{Q}^*(s, \Tilde{\pi}_\beta(s)) - \hat{Q}^*(s, \pi_\beta(s))\right]
-
2 (\Tilde{\alpha} + \alpha) \overline{\rho}_{\pi_\beta}. \quad \text{(\cref{eq: upper bound of delta difference})} \label{eq: after alpha}
\end{align}

Last, we consider the value range of $\overline{\rho}_{\pi_\beta}$.
Considering an extreme case that $\pi_\beta$ gets stuck in some state $s^\prime$, \textit{i.e.} $\forall t, P(s_t = s^\prime) = 1$, then $\overline{\rho}_{\pi_\beta} = \sum_{t=0}^\infty \gamma^t = \frac{1}{1-\gamma}$. This is obviously the upper bound of  $\overline{\rho}_{\pi_\beta}$.
On the other hand, considering another extreme case that $\pi_\beta$ visits all the states in the offline dataset $\mathcal{D}$ equiprobably, \textit{i.e.}  $\forall t, \forall s^\prime \in \mathcal{D}, P(s_t = s^\prime) = \frac{1}{|\mathcal{S}_{\mathcal{D}}|}$, then $\forall s \in \mathcal{S}_{\mathcal{D}}, \rho_{\pi_\beta}(s) = \sum_{t=0}^\infty \gamma^t \frac{1}{|\mathcal{S}_{\mathcal{D}}|} = \frac{1}{|\mathcal{S}_{\mathcal{D}}|(1-\gamma)}$.
It is impossible for any $\overline{\rho}_{\pi_\beta}$ to be less than $\frac{1}{|\mathcal{S}_{\mathcal{D}}|(1-\gamma)}$ (otherwise $\int_{s\in \mathcal{S}_{\mathcal{D}}}\overline{\rho}_{\pi_\beta} \diff s <  \int_{s\in \mathcal{S}_{\mathcal{D}}} \frac{1}{|\mathcal{S}_{\mathcal{D}}|(1-\gamma)} \diff s = \frac{1}{1-\gamma} = \int_{s\in \mathcal{S}_{\mathcal{D}}} \rho_{\pi_\beta}(s) \diff s$, which contradicts $\forall s, \overline{\rho}_{\pi_\beta} \ge \rho_{\pi_\beta}(s)$).
Therefore, $\overline{\rho}_{\pi_\beta} \in \left[\frac{1}{|\mathcal{S}_{\mathcal{D}}|(1-\gamma)}, \frac{1}{1-\gamma}\right]$.

Thus, \cref{prop: imperfect case} is proven.   \textbf{Q.E.D.}

\section{Implementation Details}
\label{appendix: Implementation Details}
\textbf{Blackbox Policy.} To play the role of proprietary preference models in real-world deployment, blackbox policies that provide action preferences are pre-trained using the SOTA algorithms and unlimited training resources in this paper.
For Gym and Adroit tasks, we adopt the online training scheme and train the SAC~\cite{sac} algorithm for 3Mil steps, following D4RL~\cite{Fu2020D4RLDF} and the rlkit\footnote[4]{\url{https://github.com/rail-berkeley/rlkit}, commit ID c81509d982b4d52a6239e7bfe7d2540e3d3cd986.} repository.
For AntMaze tasks, online algorithms fail, and we adopt the Offline-to-Online training scheme with the IQL~\cite{IQL} method, following its author-provided implementations (\textit{i.e.}, 1Mil steps for offline pre-training and 1Mil steps for online fine-tuning). 
\cref{table: appendix expert policy} shows the performance of well-trained blackbox policies over 100 random rollouts.

\begin{table}[ht]
  \small
  \caption{The training schemes and normalized D4RL scores of adopted blackbox policies.
  }
  \centering
  \begin{tabular}{|l|cccc|cccccc|}
    \hline
	Dataset
	&  HC & Hop & W
	& Pen & AM-u & AM-ud & AM-ml & AM-md & AM-lp &AM-ld  
	\\ \hline
    Training Scheme 
    & \multicolumn{4}{c|}{Online}    & \multicolumn{6}{c|}{Offline-to-Online} 
    \\ \hline
    Normalized D4RL Score 
    & 118 & 110 & 105 & 125 & 91 & 78 & 85 & 83 & 47 & 34
    \\ \hline
  \end{tabular}
  \label{table: appendix expert policy}
\end{table}

\textbf{Trained Policy.}
We train the policy for 1Mil steps in the Offline, Online, and Online-Mix schemes, and additional 100k steps in the Offline-to-Online and OAP schemes.
Other technical details of the Offline-to-Online scheme follow the AWAC and IQL~\cite{AWAC,IQL} methods.
Based on the three parts in \cref{sec: method}, the pseudo-code describing the entire training process of OAP is presented in \cref{alg: algos-OAP}.
The hyperparameters of OAP instantiated on TD3+BC~\cite{td3+bc} and IQL~\cite{IQL} are presented in \cref{table: appendix td3_hyp}.

\section{Potential Negative Societal Impact}
\label{appendix: Potential Negative Societal Impact}
RL agents may take suboptimal or even unreasonable actions during the trial-and-error training process.
Meanwhile, online interactions with the environment can be high-risk and high-cost in real-world applications, such as autonomous driving and medical treatment.
Hence, offline RL provides a more feasible solution than online RL by leveraging the offline logged data to dispense with online interactions during the training phase.
However, a limitation of offline RL is that agents' performances are primarily affected by the quality and quantity of previously collected data.
Moreover, this may include potentially damaging applications such as biased datasets and biased~agents.
For the proposed Offline-with-Action-Preferences (OAP) method, preference learning is involved in offline reinforcement learning.
We foresee the impact of our work is probably to help explore user-adaptive RL agents.
However, this characteristic may facilitate harmful applications like biased agents at the same time.
Therefore, we advocate that RL-based robotics systems, game AI and other applications should follow fair and safe principles.

\begin{table}[ht]
\vspace{-1pt}
\caption{Hyperparameters of OAP instantiated on the TD3+BC~\cite{td3+bc} and IQL~\cite{IQL} algorithms on Gym/ AntMaze/ Adroit domains. ``Unqueried First'' means selecting unqueried samples preferentially.}
\centering
\begin{tabular}{cll}
\toprule
& Hyperparameter & Value \\
\midrule
\multirow{9}{*}{TD3 Hyperparameters} & Optimizer & Adam~\cite{adam} \\
                                      & Critic learning rate & 3e-4 \\
                                      & Actor learning rate  & 3e-4 \\
                                      & Mini-batch size      & 256 \\
                                      & Discount factor      & 0.99 \\
                                      & Target update rate   & 5e-3 \\
                                      & Policy noise         & 0.2 \\
                                      & Policy noise clipping & (-0.5, 0.5) \\
                                      & Policy update frequency & 2 \\
\midrule
\multirow{6}{*}{TD3 Architecture}     & Critic hidden dim    & 256        \\
                                      & Critic hidden layers & 2          \\
                                      & Critic activation function & ReLU~\cite{relu} \\
                                      & Actor hidden dim     & 256        \\
                                      & Actor hidden layers  & 2          \\
                                      & Actor activation function & ReLU~\cite{relu} \\
\midrule
\multirow{2}{*}{TD3+BC Hyperparameters}  & $\lambda$             & 2.5 / 2.5 / 0.1 \\
                                         & State normalization            & True\\
\midrule
\multirow{6}{*}{IQL Hyperparameters}  & Optimizer & Adam~\cite{adam} \\
                                      & Policy learning rate & 3e-4 \\
                                      & Mini-batch size      & 256 \\
                                      & Dropout rate         & 0.0 / 0.0 / 0.1 \\
                                      & Beta        & 3 / 10 / 0.5 \\
                                      & Quantile & 0.7 / 0.9 / 0.7 \\
\midrule
\multirow{5}{*}{OAP Hyperparameters}    
                                      &  Training steps   & 1e6      \\
                                      &  Query limit   & 1e5      \\
                                      &  Periodical steps  & 1e5       \\
                                      &  Unqueried First  & True       \\    
                                      & L2R training epochs      & 100    \\
\midrule
\multirow{4}{*}{RankNet Architecture}            
                                      & Hidden dim          & 512, 256    \\
                                      & Hidden layers       & 2      \\
                                      & Dropout Rate        & 0.5 \\  
                                      & Activation function & ReLU~\cite{relu} \\
\bottomrule
\end{tabular}
\label{table: appendix td3_hyp}
\end{table}

\end{document}